\documentclass[final,5p,times,twocolumn]{elsarticle}

\usepackage{wrapfig}
\usepackage{flushend}
\usepackage{amssymb}
\usepackage{amsmath}
\usepackage{fontawesome}
\usepackage{float}
\usepackage{booktabs}
\usepackage{graphicx}
\usepackage{tabularx}
\usepackage{color}
\usepackage{multirow}
\usepackage[switch]{lineno} 

\def\ie{\emph{i.e., }}
\def\eg{\emph{e.g., }}

\journal{Elsevier}

\begin{document}

\begin{frontmatter}



\title{ToCoAD: Two-Stage Contrastive Learning for Industrial Anomaly Detection}


\author[1]{Yun Liang} 
\ead{yliang@scau.edu.cn}
\author[1]{Zhiguang Hu}
\ead{huzhiguang2000@stu.scau.edu.cn}
\author[1]{Junjie Huang}
\ead{junjie666@stu.scau.edu.cn}
\author[2]{Donglin Di}
\ead{didonglin@lixiang.com}
\author[3]{Anyang Su}
\ead{aysu17@mails.jlu.edu.cn}
\author[4]{Lei Fan\corref{corresponding}}
\ead{lei.fan1@unsw.edu.au}

\cortext[corresponding]{Corresponding author.}

\affiliation[1]{organization={College of Mathematics and Informatics},
            addressline={South China Agricultural University}, 
            city={Guangzhou},
            postcode={510642}, 
            country={China}}
\affiliation[2]{organization={Space AI},
            addressline={Li Auto}, 
            city={Beijing},
            postcode={101399}, 
            country={China}}
\affiliation[3]{organization={College of Software},
            addressline={Jilin University}, 
            city={Changchun},
            postcode={130012}, 
            country={China}}
\affiliation[4]{organization={School of Computer Science and Engineering},
            addressline={The University of New South Wales}, 
            city={Sydney},
            postcode={2052}, 
            country={Australia}}

\begin{abstract}
Current unsupervised anomaly detection approaches perform well on public datasets but struggle with specific anomaly types due to the domain gap between pre-trained feature extractors and target-specific domains. To tackle this issue, this paper presents a two-stage training strategy, called \textbf{ToCoAD}. In the first stage, a discriminative network is trained by using synthetic anomalies in a self-supervised learning manner. This network is then utilized in the second stage to provide a negative feature guide, aiding in the training of the feature extractor through bootstrap contrastive learning. This approach enables the model to progressively learn the distribution of anomalies specific to industrial datasets, effectively enhancing its generalizability to various types of anomalies. Extensive experiments are conducted to demonstrate the effectiveness of our proposed two-stage training strategy, and our model produces competitive performance, achieving pixel-level AUROC scores of 98.21\%, 98.43\% and 97.70\% on MVTec AD, VisA and BTAD respectively.
\end{abstract}




\end{frontmatter}




\section{Introduction}

Industrial anomaly detection aims to identify defective products during quality inspection sessions within the production process, intending to improve the yield rate. Recently, interest in this field has surged due to the growing need of industrial development~\cite{tao2022survey22, liu2024survey2023}. Due to the challenge of obtaining sufficient anomalous samples, the distribution of anomalies is non-estimable, making this scenario typically classified as an unsupervised learning task. In such unsupervised settings, the objective is to train models using only defect-free samples, enabling them to detect and localize anomalous regions during the testing phase.

\begin{figure}
    \centering
    \includegraphics[width=\linewidth]{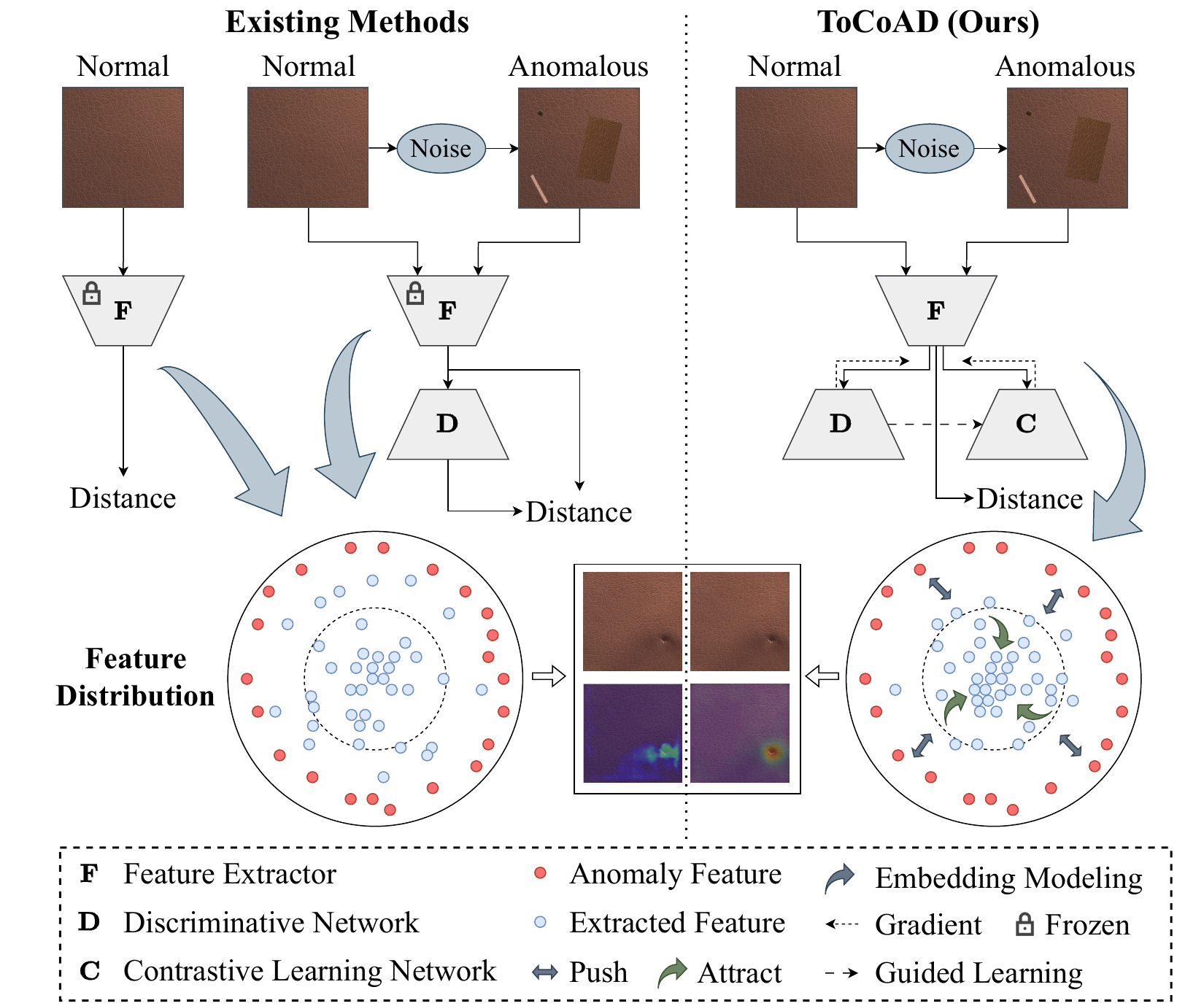}
    \caption{
    Existing methods rely on frozen pre-trained feature extractors, which can lead to inaccuracies in anomaly detection. In contrast, our method utilizes a two-stage training strategy to fine-tune the feature extractor under the contrastive learning paradigm.
    }
    \label{fig:compare}
\end{figure}

Existing anomaly detection methods, predominantly leveraging deep learning techniques, have shown superior performance in industrial benchmarks~\cite{bergmann2019mvtec, batzner2024efficientad}. As illustrated in Figure~\ref{fig:compare}, the majority of existing methods, including feature embedding-based~\cite{defard2021padim, roth2022towards,cohen2020spade,gudovskiy2022cflowad} and synthetic anomaly-based methods~\cite{zavrtanik2021draem, schluter2022nsa, li2021cutpaste}, have attempted to train a classifier or discriminative network to identify anomalies by utilizing a pre-trained truncated model for extracting features. However, it is challenging to accurately identify all anomalies, primarily due to the indistinguishability of frozen pre-trained models from normal and anomalous features. This issue arises from the domain gaps~\cite{you2019uda,nam2021reducingdomaingap} between pre-trained datasets (\eg ImageNet~\cite{deng2009imagenet}) and target-specific domains (\ie industrial images). Recently, reconstruction-based methods~\cite{zavrtanik2021riad, pirnay2022intra, wyatt2022anoddpm} have been developed to train a decoder and feature extractor jointly by learning an identical map between the input and output. However, the performance of these methods remains suboptimal, limited by the training overhead and the ability to reconstruct large areas of anomalies. 

Recently, the application of contrastive learning has demonstrated considerable progress in cross-domain learning~\cite{huang2022caco, thota2021cda}. 
High-quality image classification~\cite{dao2023multilabelimageclassification}, segmentation~\cite{ki2021weaklysupervisedobjectlocalization, wang2023dsfwsi, wang2023oneshotorgansegmentation}, or deraining~\cite{chen2022unpairedderain} is achieved by introducing several branches of contrastive learning methods, which can minimize the semantic gaps between target datasets and pre-trained datasets. 
Fine-tuning the network using positive sample pairs or negative samples has been demonstrated to be effective in enhancing the network's generalizability to domain-specific datasets, indicating significant potential in the field of anomaly detection as well. 

To tackle these challenges, we propose a two-stage training strategy that incorporates contrastive learning due to its advantages in learning robust features from target samples through self-supervised learning~\cite{tian2020explorecl}. Specifically, we initially leverage a frozen feature extractor to obtain generalized features and train a discriminative network progressively to identify anomalies using synthetic anomalous samples. In the second stage, the discriminate network is fixed to provide a negative guide, while a bootstrap contrastive learning is employed to fine-tune the feature extractor and contrastive learning network jointly. This joint learning can be viewed as a form of adversarial learning~\cite{kim2020advsslcl, ho2020clae} of defect and defect-free features, where normal features are compactly enclosed and distinct from defective features. Finally, the fine-tuned feature extractor is used to extract patch features to construct a memory bank through coreset subsampling, and distance metrics between the test sample features and the memory bank features are computed to localize anomalies during the inference phase.

Our contributions can be summarized as follows:

\begin{itemize}
  \item We propose two-stage training strategy to fine-tune the feature extractor to bridge the domain gap between pre-trained and target features. In the first stage, a network is progressively trained to coarsely localize anomalies, and then it is employed to facilitate the fine-tuning of the extractor in the second stage. 
  \item We introduce a negative-guided contrastive learning paradigm, which utilizes the discriminative network guiding negative bootstrap contrastive learning to fine-tune the feature extractor. A joint learning contrastive loss is introduced to regularize negative bootstrap learning on the model.  
  \item Our model shows competitive performance on three popular anomaly detection datasets. It achieves AUROC scores of 99.10\% / 98.21\% (image / pixel-level) on MVTec AD, 95.35\% / 98.43\% on VisA, and 97.70\% (pixel-level) on BTAD.
\end{itemize}
\begin{figure*}
    \centering
    \includegraphics[width=\linewidth]{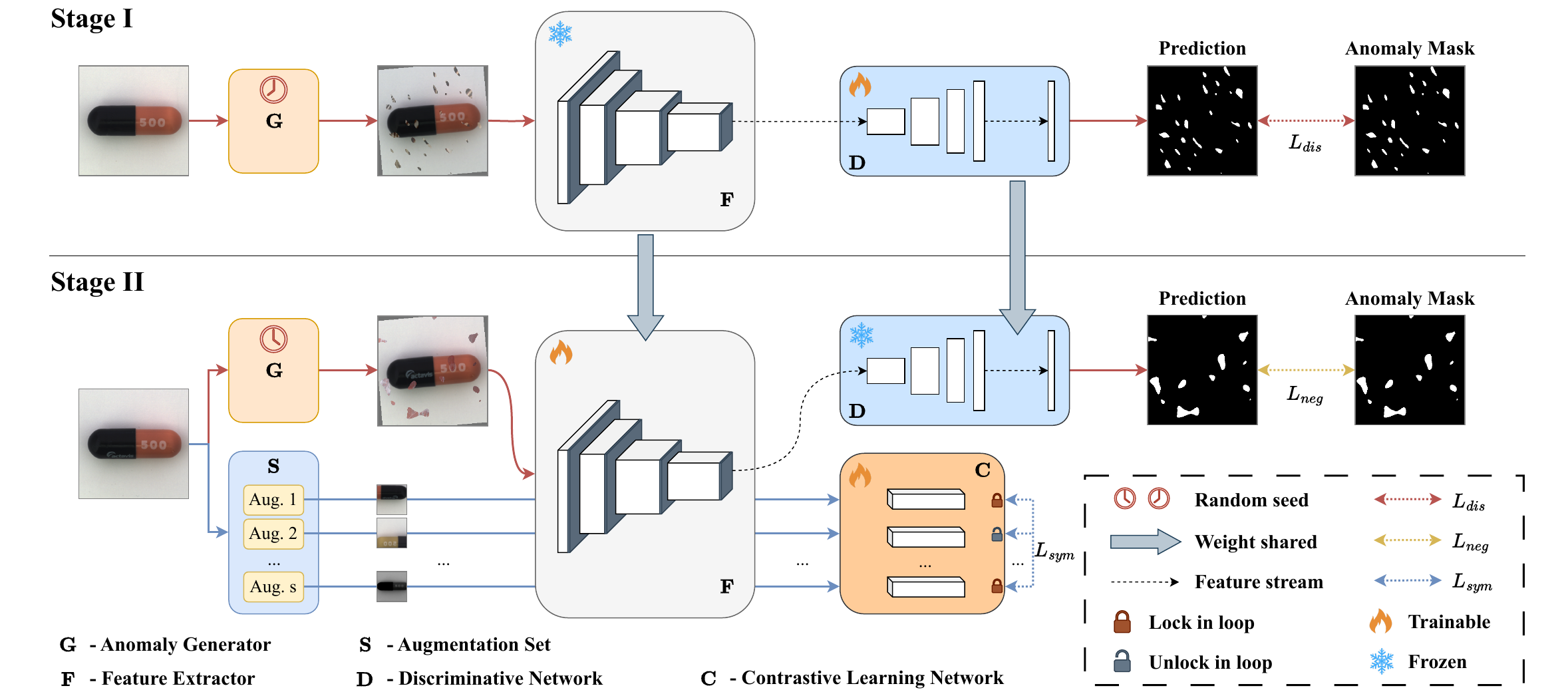}
    \caption{
    Overview of our two-stage training strategy, ToCoAD. First, a synthetic anomaly image is generated by anomaly generator $\mathbf{G}$ to train a discriminative network $\mathbf{D}$. Then, the feature extractor $\mathbf{F}$ is fune-tuned jointly by contrastive learning network $\mathbf{C}$ and pre-trained discriminative networks $\mathbf{D}$ using synthetic anomaly images and augmented image pairs.
    }
    \label{fig:trainpipeline}
\end{figure*}

\section{Related Work}
\subsection{Anomaly Detection}
Most anomaly detection methods utilize a feature extractor pre-trained on ImageNet to obtain generalized features. These methods can be classified into three types: feature embedding-based, reconstruction-based and synthetic anomaly-based methods.

\textbf{Feature embedding-based methods}: They typically utilize pre-trained models to extract features and directly model the distributions using various machine learning techniques. For example, SPADE~\cite{cohen2020spade} uses the K-Nearest-Neighbors (KNN) algorithm~\cite{cover1967knn} to obtain representative features extracted by a pre-trained ResNet~\cite{he2016resnet}. PaDiM~\cite{defard2021padim} decomposes the image into patches to obtain a probabilistic representation of the normal class using multivariate Gaussian distributions. PatchCore~\cite{roth2022towards} constructs a memory bank for storing neighborhood-aware patch-level features. 
Some methods are based on Normalizing Flow~\cite{rezende2015nf} (NF) to convert pre-trained feature distributions into simple distributions. CFLOW-AD~\cite{gudovskiy2022cflowad} utilizes a pre-trained encoder with multiscale pyramid pooling to capture rich features and leverages conditional normalization flows to enhance anomaly detection efficiency. PyramidFlow~\cite{lei2023pyramidflow} utilizes invertible pyramids and coupled pyramid blocks to localize anomalies through multi-scale feature interaction. SANF~\cite{ma2024sanf} uses a pre-trained Vision Transformer (ViT)~\cite{dosovitskiy2020vit} to extract semantic and spatial features from an image for feature fusion. 
These methods use pre-trained feature extractors, such as ResNet~\cite{he2016resnet}, WideResNet~\cite{zagoruyko2016wrn}, EfficientNet~\cite{tan2019efficientnet} and ViT~\cite{dosovitskiy2020vit} to extract features for modeling distributions. 

\textbf{Reconstruction-based methods}: They assume that models trained on only normal samples cannot accurately reconstruct anomalous regions, thus allowing for the localization of anomalies by identifying differences between the input and reconstruction results. For instance, RIAD~\cite{zavrtanik2021riad} and InTra~\cite{pirnay2022intra} perform mask operations on normal samples, and train reconstruction models to recover the masked regions. DiffusionAD~\cite{zhang2023diffusionad} uses a diffusion model~\cite{ho2020ddpm} to reconstruct normal samples as near-normal samples, and localizes anomalous regions using a segmentation network.

\textbf{Synthetic anomaly-based methods}: They formulate unsupervised anomaly detection as a binary classification task, in which pseudo anomaly samples are generated to train the discriminative network for identifying anomalies. CutPaste ~\cite{li2021cutpaste} employs an augmentation strategy where a smaller patch is replaced by other regions. MemSeg~\cite{yang2023memseg} and DRAEM~\cite{zavrtanik2021draem} combine Perlin noise~\cite{perlin1985perlin} and binarized masks of samples to generate anomalous images for training their discriminative networks. NSA~\cite{schluter2022nsa} employs poisson image editing techniques to achieve seamless fusion of anomalies. 

However, most of these methods rely on feature extractors pre-trained on ImageNet, leading to a domain gap between pre-trained features and industrial target features. To alleviate this problem, our ToCoAD employs a two-stage training approach that uses positive and negative samples to jointly fine-tune the feature extractor, guiding it to acquire adaptive feature representation capabilities.

\subsection{Contrastive Learning}
Recently, contrastive learning~\cite{chen2021simsiam, grill2020byol, he2020moco, cover1967knn} has played a significant role in self-supervised learning, offering a promising paradigm for exploiting unlabeled data without the need for human annotations. The concept aims to enhance feature consistency and obtain invariant feature representations between different views of the same sample, while preserving the differences among other samples. For example, SimCLR~\cite{chen2020simclr} and MoCo~\cite{he2020moco} construct positive pairs from the same sample and negative pairs from different samples, employing an N-pair loss~\cite{sohn2016npairloss} to keep positive pairs close and negative pairs far away. BYOL~\cite{grill2020byol} avoids prediction crashes and removes negative pairs by an exponential moving average model. SWaV~\cite{caron2020swav} compares cluster assignments under different views instead of directly comparing features for self-supervised learning. SimSiam~\cite{chen2021simsiam} addresses the issue of collapsing solutions~\cite{zhang2022whysimsiam} without using negative samples by employing a stop-gradient operation and a predictor.

Given the capability of contrastive learning methods to enable models to acquire robust and generalized feature representations~\cite{zhang2022dac,thota2021cda,wang2022cdcl,huang2022caco,wang2023dsfwsi}, we propose a negative bootstrap contrastive learning to fine-tune the feature extractor. Our approach uses augmented positive samples to train the model by learning feature representations of normal samples while simultaneously using synthetic anomaly samples to bootstrap the model, ensuring its sensitivity to anomalous features.
\section{Method}
Given a training set \(\mathcal{T}_{\text{train}}\) comprising $N$ images and a test set \(\mathcal{T}_{\text{test}}\) consisting of $N^\prime$ images, each image $I\in \mathbb{R}^{H\times W \times C}$ corresponds to a binary label $c=\{0, 1\}$ where $H$, $W$ and $C$ denote the height, width and the number of the channel of the image. For unsupervised anomaly detection, all images in \(\mathcal{T}_{\text{train}}\) are categorized into normal (\eg defect-free) images with $c=0$, while \(\mathcal{T}_{\text{test}}\) may contain anomalous images with $c=1$ and a corresponding binary annotation $y\in \{0, 1\}^{H\times W\times C}$ indicating the anomalous regions. The normal images in the training set and the test set come from the same normal data distribution with similar sample characteristics, while the anomalous images deviate from these normal samples. The goal is to classify and localize anomalous regions in \(\mathcal{T}_{\text{test}}\). 

As shown in Figure~\ref{fig:trainpipeline}, we propose a two-stage training strategy to learn a feature representation adapted to the target data distribution. The first stage is to train a discriminative network to detect anomalies and localize anomalous regions coarsely (Section~\ref{sec:DNP}). It includes an anomaly generator $\mathbf{G}$, a feature extractor $\mathbf{F}$ and a discriminative network $\mathbf{D}$. In the second stage, the discriminative network pre-trained in stage I is incorporated with the contrastive learning network $\mathbf{C}$, and these networks are jointly trained in a bootstrap contrast learning manner (Section~\ref{sec:NCL}). Finally, a memory bank $\mathcal{M}$ is constructed for storing the normal features, which is used to estimate the anomaly score maps during the testing phase (Section~\ref{sec:MMI}).

\subsection{Discriminative Network Pre-training}
\label{sec:DNP}
Given the unavailability of anomalous samples for training a discriminative network, we initially utilize an anomaly generator $\mathbf{G}$ to synthesize pseudo anomalies from the training set. The generation of anomalies can be implemented in various forms by using self-supervised learning techniques~\cite{li2021cutpaste, schluter2022nsa} or by injecting noises into normal images~\cite{zavrtanik2021draem, liu2023simplenet, yang2023memseg}. 
 To obtain discriminative representations of various artificial defects, we employ widely-used Perlin noise~\cite{perlin1985perlin} as the anomaly generator $\mathbf{G}$, which synthesizes pseudo-anomaly samples by injecting Perlin noise into normal images.

\begin{figure}
    \centering
    \includegraphics[width=\linewidth]{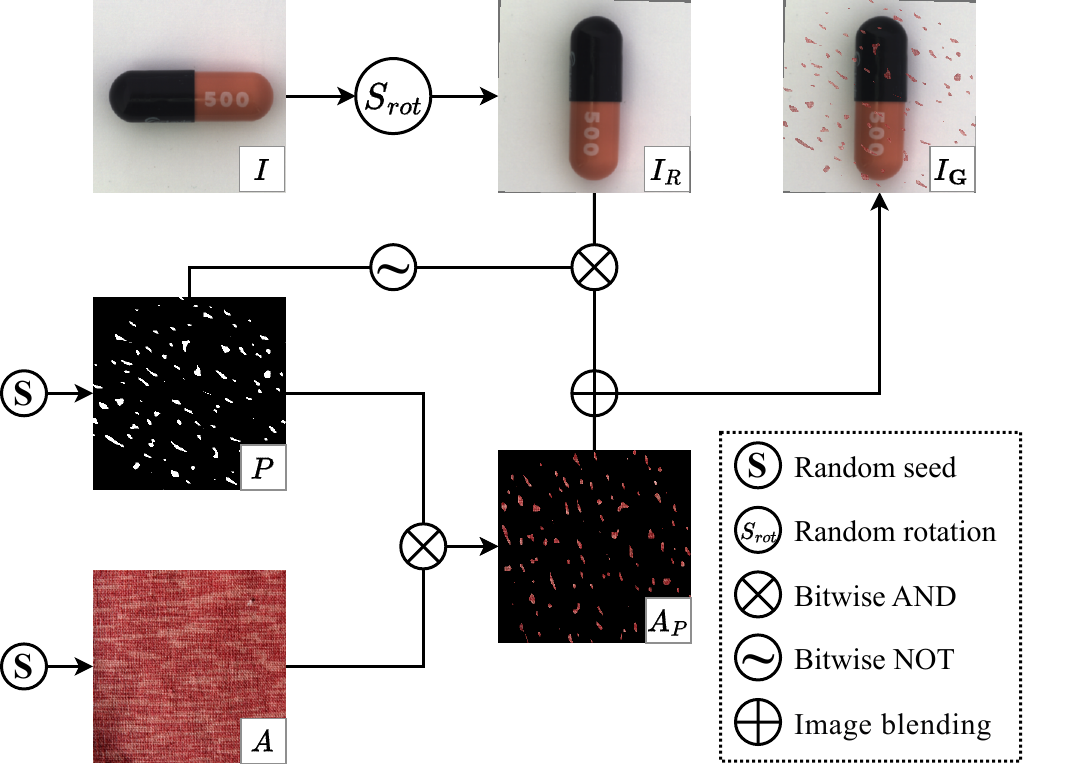}
    \caption{
    Overview of anomaly generator $\mathbf{G}$ with Perlin noise. Firstly, the normal image $I$ undergoes random slight angle and 90-degree rotation to obtain $I_R$. Secondly, the Perlin noise $P$ and the texture sample $A$ are subjected to a bitwise-and operation to generate the anomalous region $A_P$. Finally, the anomalous region $A_P$ and the rotated image $I_R$ are fused to obtain the synthetic anomaly sample $I_\mathbf{G}$.
    }
    \label{fig:anomaly_generator}
\end{figure}
 
 As shown in Figure~\ref{fig:anomaly_generator}, random slight angle rotations and random \{0, 90, 180, 270\} degree rotations are applied to a base (normal) image, thereby increasing the diversity of the normal sample. Then, a random seed is employed to generate Perlin noise, with the threshold value subsequently adjusted to control the size, quantity and position of the Perlin noise. An additional texture dataset, such as the Describable Textures Dataset~\cite{cimpoi2014dtd}, is introduced as references to combine with normal samples based on the Perlin noise mask, which are considered as synthesized anomalous regions. Finally, the anomalous regions are pasted onto the base sample image to obtain the synthetic anomaly image and its corresponding mask.
 A normal image $I \in \mathcal{T}_{\text{train}}$ is passed through the anomaly generator $\mathbf{G}$ to produce a pseudo-anomaly sample $I_{\mathbf{G}}$ with an anomaly mask $y_{\mathbf{G}} \in \{0,1\}^{H \times W \times C}$. A pseudo-anomaly dataset can be obtained:

\begin{equation}
    \mathcal{T}_\mathbf{G}=\mathbf{G}(I_i),\ \forall\  I_i\in\mathcal{T}_{\text{train}},
\end{equation}
where $i=\{1,\dots, N\}$ denotes the index in the training set.

Then, each pseudo-anomaly image is used to train the discriminative network $\mathbf{D}$, where the pre-trained feature extractor is fixed during the early stages of model training. The architecture of the discriminative network is symmetric with the feature extractor, such as the WideResNet50~\cite{zagoruyko2016wrn}. This network operates by successively up-sampling and splicing the extracted features to finally output a feature map \(\hat{y}_{\mathbf{G}}\) as a predicted mask with the same dimensions as the input image $I_{\mathbf{G}}$. The training loss $L_{dis}$ is computed as:

\begin{equation}
    \label{eq:focalloss}
    L_{dis}=L_{focal}(\hat{y}_{\mathbf{G}}, y_{\mathbf{G}})=-\alpha_t\left(1-p_t\right)^\gamma \log \left(p_t\right),
\end{equation}
where \(\alpha_t\) is the scaling factor associated with the category \(t\), \(\gamma\) is an adjustable parameter, and \(p_t\) corresponds to the predicted pixel point classification with 1 for the anomalous category and 0 for the normal category. The Focal Loss~\cite{lin2017focalloss} effectively addresses the sample imbalance problem in pixel-level one-class classification. Upon completion of the discriminative network training, the extracted feature map can coarsely localize the probability of anomalies in the input image.

\subsection{Negative-guided Contrastive Learning}
\label{sec:NCL}
\begin{figure}
    \centering
    \includegraphics[width=\linewidth]{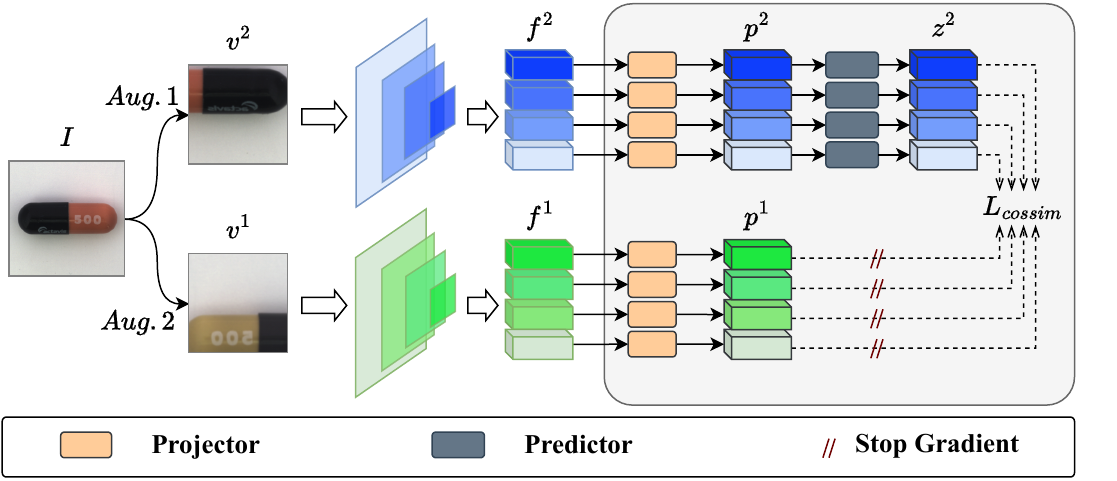}
    \caption{
    Detailed structure of contrastive learning network $\mathbf{C}$. A normal image $I$ is augmented to obtain two views $v^1,v^2$, which are then passed through the feature extractor and a contrastive learning network to obtain the projected features $z^1,z^2$ and predicted features $p^1,p^2$. The $L_{cossim}$ is calculated for two branches, one of which has no predictor branch applying the stop-gradient operation.
    }
    \label{fig:contrastivenetwork}
\end{figure}

In stage II, the discriminative network $\mathbf{D}$ is employed to train the feature extractor $\mathbf{F}$ and the contrastive learning network \(\mathbf{C}\). Specifically, discriminative network $\mathbf{D}$, pre-trained in stage I, acts as a guide for providing negative sample features in bootstrap learning. We employ the advanced SimSiam~\cite{chen2021simsiam} as the contrastive learning network $\mathbf{C}$ due to its simplicity and exceptional capability in achieving robust and generalized feature representation.
As shown in Figure~\ref{fig:contrastivenetwork}, for training the contrastive learning network $\mathbf{C}$, a normal image $I$ is augmented using a data augmentation set $\mathbf{S}$ (such as random crop, color jitter, grayscale) to generate $M$ different views $v^1,\dots,v^M$. These views are fed into the feature extractors $\mathbf{F}$ to extract features from different layers, which are considered as generic feature representations from a normal image, and these features contain both the low-level texture detail feature representation and the high-level abstract semantic feature representation. 

For constructing contrastive instance pairs, we take an input image $I$ with two augmented views $v^1$ and $v^2$ ($M$ set 2). As an example, the features $f^1$ are extracted from $v^1$ using feature extractor $\mathbf{F}$ and then are passed into the projector and predictor sequentially to obtain $z_1$ and $p_1$ respectively. 
Then, the cosine similarity loss can be obtained:

\begin{equation}
    \label{eq:cosimloss}
    \mathcal{D}\left(p_1, z_2\right)=-\frac{p_1}{\left\|p_1\right\|_2} \cdot \frac{z_2}{\left\|z_2\right\|_2}, 
\end{equation}
where \(\left\|\cdot\right\|_2\) denotes the L2 distance of an output feature. Following previous work~\cite{chen2021simsiam, wang2023dsfwsi}, the stop-gradient (SG) operation is employed to halt the gradient propagation of $z_1$ and $z_2$, effectively preventing collapsing solutions. Since \(\mathcal{D}\) is asymmetric, the average bidirectional similarity between $f^1$ and $f^2$ is computed to maintain equilibrium, as shown below:

\begin{equation}
    \label{eq:symcosimloss}
    L_{cossim}(f^1,f^2)=\frac{1}{2} \mathcal{D}\left(p_1, \text{SG}(z_2)\right)+\frac{1}{2} \mathcal{D}\left(p_2, \text{SG}(z_1)\right),
\end{equation}

When generating \(M\) augmented samples from each normal sample, the symmetric cosine similarity loss is computed as:

\begin{equation}
\begin{aligned}
    \label{eq:totalsymcosimloss}
    L_{sym}&=\frac{2}{M(M-1)} \sum_{i}^{M} \sum_{j \neq i}^{M} L_{cossim}(f^i,f^j) \\
    &=\sum_{i}^{M} \sum_{j \neq i}^{M}\frac{\left[\mathcal{D}\left(p_{j}, \operatorname{SG}\left(z_{i}\right)\right)+\mathcal{D}\left(p_{i}, \operatorname{SG}\left(z_{j}\right)\right)\right]}{M(M-1)}.
\end{aligned}
\end{equation}

For synthetic anomaly sample, we obtain its pseudo-anomaly mask \(y^\prime_{\mathbf{G}}\) and predictive feature map \(\hat{y}^\prime_{\mathbf{G}}\), and compute the Focal Loss \(L_{neg}=L_{focal}(\hat{y}^\prime_{\mathbf{G}}, y^\prime_{\mathbf{G}})\) as Equation~\ref{eq:focalloss}. The training loss in the second training stage is shown as:
\begin{equation}
    \label{eq:2ndloss}
    L_{ncl}=\lambda\cdot L_{sym}+(1-\lambda)\cdot L_{neg},
\end{equation}
where \(\lambda\) is a hyperparameter used to balance these two losses.

\subsection{Memory Modeling and Anomaly Detection}
\label{sec:MMI}
\begin{figure}
    \centering
    \includegraphics[width=\linewidth]{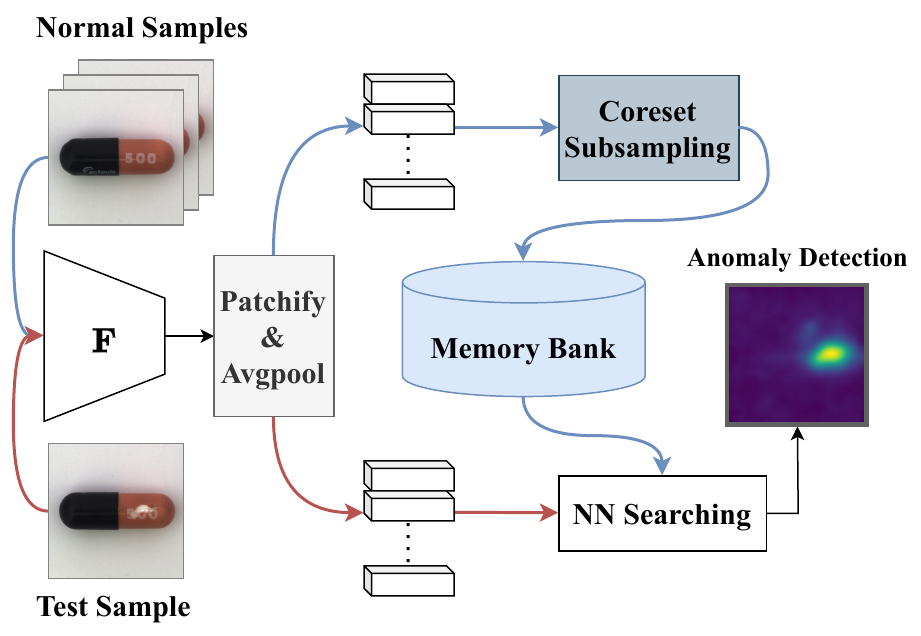}
    \caption{
    Pipeline of modeling memory bank. The fine-tuned feature extractor is used to extract the adapted patch features from normal samples, and then these features are stored in the memory bank through coreset subsampling. During the inference phase, anomalies are detected by calculating the Euclidean distance between the adapted features from the test image and the nearest neighbor coresets.
    }
    \label{fig:memory}
\end{figure}

During the inference phase, we utilize the feature extractor as a backbone to extract adapted features from various layers. These features are then compressed into a memory bank via the coreset selection~\cite{roth2022towards}, as shown in Figure~\ref{fig:memory}.
To concurrently harvest shallow texture information and deep semantic content, features extracted from both layer 2 and layer 3 are aggregated to derive a comprehensive feature for each image. These features are then used to construct the original memory bank $\mathcal{M}_O$. To minimize storage and computation overhead, we use the greedy coreset sampling algorithm to identify the most representative features. This results in a sampled memory bank $\mathcal{M}$ obtained by solving Equation~\ref{eq:nphard} using iterative greedy approximation suggested in~\cite{sener2017coreset}.

\begin{equation}
\label{eq:nphard}
\mathcal{M}^*=\underset{\mathcal{M} \subset \mathcal{M}_O}{\arg \min } \max _{m \in \mathcal{M}_O} \min _{n \in \mathcal{M}}\|m-n\|_2.
\end{equation}

When testing an input image \(I_t\in\mathcal{T}_{\text{test}}\), pixel-level anomaly score \(s_t\) is calculated by maximum Euclidean distance between its adapt patch features \(p_t\) and its nearest normal adapted features coreset \(c^*\) from memory bank \(\mathcal{M}\):

\begin{equation}
s_t^{\prime}=\min _{c^* \in \mathcal{M}} E\left(p_t, c^*\right),
\end{equation}
\begin{equation}
s_t=\left(1-\frac{\exp(s_t^{\prime})}{\sum_{c^{\prime} \in \mathcal{N}_b\left(c^*\right)} \exp(E\left(p_t, c^{\prime}\right))}\right) s_t^{\prime}.
\end{equation}
where \(E(\cdot)\) is Euclidean distance and \(\mathcal{N}_b\) is the b-nearest neighbor coresets of \(c^*\) in the memory bank. Finally, the image-level anomaly score is calculated by the maximum anomaly score for all patches in the image.

\begin{table*}[ht]
  \caption{Image-level and pixel-level AUROC (\%) on MVTec AD dataset. Average results are reported in 5 texture categories, 10 texture categories, and all categories, respectively. The best results are shown in bold.}
  \label{tab:mvtecad}
  {\fontsize{7}{14}\selectfont
  \setlength{\tabcolsep}{4pt}
  \begin{center}
  \begin{tabularx}{\textwidth}{l|ccccccccc|cc}
    \hline
      & SPADE & 
        PaDiM & 
        PatchCore & 
        FAPM & 
        RD4AD & 
        CFLOW-AD & 
        DeSTSeg & 
        PyramidFlow & 
        MMR & 
        ToCoAD &
        ToCoAD \\
        & \cite{cohen2020spade} & \cite{defard2021padim} & \cite{roth2022towards} & \cite{kim2023fapm} & 
        \cite{deng2022rd4ad} & 
        \cite{gudovskiy2022cflowad} & \cite{zhang2023destseg} & \cite{lei2023pyramidflow} & 
        \cite{zhang2023mmr} & 
        + CutPaste~\cite{li2021cutpaste} &
        + Perlin~\cite{perlin1985perlin}
        \\
    \hline
    Bottle & 98.1 / 98.4 & - / 98.3 & \textbf{100} / \textbf{98.6} & \textbf{100} / 98.2 & 98.7 / 96.6 & 98.9 / 96.8 & - / \textbf{99.2} & - / 95.9 & \textbf{100} / 98.3 & \textbf{100} / 98.4 & \textbf{100} / 98.5\\
    
    Cable& 93.2 / 97.1 & - / 96.7 & 99.5 / 98.4 & \textbf{99.2} / \textbf{98.5} & 97.4 / 91.0 & 97.6 / 93.5 & - / 97.3 & - / 92.1 & 97.8 / 95.4 & 98.9 / 98.2 & 98.7 / 98.3\\
    
    Capsule& 98.2 / 99.0 & - / 98.5 & 98.1 / 98.8 & 98.6 / 99.0 & 98.7 / 95.8 & \textbf{98.8} / 93.4 & - / \textbf{99.1} & - / 96.1 & 96.9 / 98.0 & 98.6 / 98.9 & 98.7 / 99.0\\
    
    Carpet& 98.5 / 97.5 & - / 99.1 & 98.4 / \textbf{99.0} & 99.0 / 98.9 & 98.9 / 98.9 & 99.0 / 97.7 & - / 96.1 & - / 90.8 & \textbf{99.6} / 98.8 & 97.9 / 98.8 & 98.0 / 98.7\\
    
    Grid& 99.0 / 93.7 & - / 97.3 & 98.2 / 98.7 & 98.0 / 97.8 & 99.3 / 97.6 & 98.7 / 96.0 & - / \textbf{99.1} & - / 94.2 & \textbf{100} / 99.0 & 98.5 / 98.4 & 98.7 / 98.5\\
    
    Hazelnut& 98.7 / 99.1 & - / 98.1 & \textbf{100} / 98.7 & 99.9 / 98.6 & 98.9 / 95.5 & 98.7 / 96.6 & - / \textbf{99.6} & - / 98.0 & \textbf{100} / 98.5 & \textbf{100} / 98.3 & \textbf{100} / 98.4\\
    
    Leather& 99.1 / 97.5 & - / 99.2 & \textbf{100} / 99.3 & 99.7 / 99.0 & 99.4 / 99.1 & 99.6 / 99.3 & - / 99.5 & - / \textbf{99.6} & \textbf{100} / 99.2 & \textbf{100} / 99.1 & \textbf{100} / 99.2\\
    
    Metal\_nut& 96.7 / 98.1 & - / 97.0 & \textbf{100} / 98.4 & \textbf{100} / 98.0 & 97.3 / 92.3 & 98.6 / 91.6 & - / \textbf{98.6} & - / 92.8 & 99.9 / 95.9 & 99.9 / 98.5 & \textbf{100} / 98.4\\
    
    Pill& 96.5 / 96.5 & - / 95.7 & 96.2 / 97.4 & 96.0 / 98.0 & 98.2 / 96.4 & \textbf{98.8} / 95.3 & - / \textbf{98.7} & - / 96.2 & 98.2 / 98.4 & 94.8 / 98.3 & 97.0 / 98.3\\
    
    Screw& 99.5 / 98.7 & - / 98.5 & 98.0 / 99.3 & 95.2 / 99.0 & \textbf{99.6} / 98.2 & 98.8 / 95.3 & - / 98.5 & - / 94.0 & 92.5 / \textbf{99.5} & 96.8 / \textbf{99.5} & 97.0 / \textbf{99.5}\\
    
    Tile& 89.8 / 87.4 & - / 94.1 & 98.4 / 95.6 & 99.4 / 95.2 & 95.6 / 90.6 & 98.0 / 94.3 & - / \textbf{98.0} & - / 97.9 & 98.7 / 95.6 & 99.2 / 97.7 & \textbf{99.5} / 97.5\\
    
    Toothbrush& 98.9 / 97.9 & - / 98.5 & 99.7 / 98.7 & \textbf{100} / 98.5 & 99.1 / 94.5 & 98.8 / 95.0 & - / \textbf{99.3} & - / 98.9 & \textbf{100} / 98.4 & \textbf{100} / 98.7 & \textbf{100} / 98.7\\
    
    Transistor& 81.0 / 94.1 & - / 97.5 & \textbf{100} / 96.3 & \textbf{100} / \textbf{98.2} & 92.5 / 78.0 & 98.0 / 81.3 & - / 89.0 & - / 97.4 & 95.1 / 90.2 & 98.4 / 94.5 & 99.8 / 95.5\\
    
    Wood& 95.0 / 88.4 & - / 94.9 & 99.2 / 95.0 & 99.1 / 94.0 & 95.3 / 90.9 & 96.6 / 95.8 & - / \textbf{97.7} & - / 93.8 & 99.1 / 94.8 & \textbf{99.8} / 95.8 & 99.6 / 95.7\\
    
    Zipper& 98.8 / 96.5 & - / 98.3 & 99.4 / 98.8 & 99.3 / 98.6 & 98.2 / 95.4 & 99.0 / 96.6 & - / \textbf{99.1} & - / 95.4 & 97.6 / 98.0 & 99.4 / 98.9 & \textbf{99.5} / 98.9\\
    
    \hline
    Texture avg.& 96.28 / 92.90 & - / 96.92 & 98.84 / 97.52 & 99.04 / 96.98 & 97.36 / 95.80 & 98.38 / 96.62 & - / \textbf{98.08} & - / 95.26 & \textbf{99.48} / 97.48 & 99.08 / 97.96 & 99.16 / 97.92\\
    Object avg.& 95.96 / 97.54 & - / 97.71 & 99.09 / 98.34 & 98.82 / 98.46 & 99.01 / \textbf{98.81} & 98.60 / 93.54 & - / 97.84 & - / 95.68 & 97.80 / 97.06 & 98.69 / 98.22 & \textbf{99.07} / 98.35\\
    
    \hline
    Total avg.& 96.06 / 95.99 & - / 97.44 & 99.01 / 98.07 & 98.89 / 97.96 & 98.46 / 97.80 & 98.22 / 94.56 & - / 97.92 & - / 95.54 & 98.36 / 97.20 & 98.82 / 98.13 & \textbf{99.10} / \textbf{98.21}\\
    
    \hline
  \end{tabularx}
  \end{center}
  }
\end{table*}

\section{Experiments}
We present the experimental settings in Section~\ref{sec:experimentsetting}, including datasets, evaluation metrics, and implementation details. Then, we demonstrate the anomaly detection efficacy of our proposed method in comparison with existing models on three datasets, and show ablation experiments in Section~\ref{sec:resultsandanalysis}.
\subsection{Experimental Settings}
\label{sec:experimentsetting}
\subsubsection{Datasets} 
We use the MVTec Anomaly Detection Dataset~\cite{bergmann2019mvtec} (MVTec AD), the Visual Anomaly (VisA) Dataset~\cite{zou2022visa} and the BeanTech Anomaly Detection Dataset~\cite{mishra2021vtadl} (BTAD) dataset for our experiments. 

\textbf{MVTec AD} is widely used as a benchmark for industrial anomaly detection. It consists of 15 categories including 5 types of texture and 10 types of object, and comprises 3,629 training images and 1,725 test images. All image sizes range from $700 \times 700$ to $1,024 \times 1,024$ pixels and each class has at least one type of defect. 

\textbf{VisA} contains 12 subsets with a total of 10,821 images, of which 9,621 are normal images and 1,200 are anomalous images. Images can be classified into three principal categories based on the intrinsic properties of the object depicted. The first category comprises single-object images typically centered in the frame. The second category encompasses multi-object. The third category includes complex printed circuit board images. 

\textbf{BTAD} consists of three industrial products, ranging in size from $600 \times 600$ to $1,600 \times 1,600$ pixels. It includes 1,799 images and 1,031 images for training and testing respectively. 

\subsubsection{Evaluation Metrics} 
Since determining of normal and anomalous samples is regarded as a binary classification problem, we use the Area Under the Receiver Operating Characteristic Curve (AUROC) metric to evaluate the anomaly detection performance. 
The ROC curve illustrates the performance of a model for binary classification across varying thresholds by delineating the correlation between the False Positive Rate (FPR) and the True Positive Rate (TPR). The AUROC is a numerical representation of the area under the ROC curve, which is used to measure the model's ability to differentiate between positive and negative categories. 
Following the previous studies~\cite{roth2022towards, defard2021padim}, to evaluate the detection performance, we calculate the image-level AUROC score between the model output anomaly scores and image-level categories. For segmentation evaluation, we measure pixel-level AUROC scores between the anomaly score map and the ground truth mask of anomalous samples.

\begin{table*}[ht]
  \caption{Image-level and pixel-level AUROC (\%) on VisA dataset. Average results are reported in 12 categories. The best results are shown in bold.}
  \label{tab:visa}
  {\fontsize{10}{14}\selectfont
  \setlength{\tabcolsep}{6pt}
  \begin{center}
  \begin{tabularx}{\textwidth}{l|ccccc|cc}
    \hline
      & SPADE & 
        DRAEM & 
        PatchCore & 
        FastFlow & 
        CFLOW-AD &   
        ToCoAD &
        ToCoAD \\
        & \cite{cohen2020spade} & \cite{zavrtanik2021draem} & \cite{roth2022towards} & 
        \cite{yu2021fastflow} & 
        \cite{gudovskiy2022cflowad} & 
        + CutPaste~\cite{li2021cutpaste} &
        + Perlin~\cite{perlin1985perlin}
        \\
    \hline
    Candle & 91.0 / 97.9 & 91.8 / 96.6 & \textbf{99.1} / 98.8 & 92.4 / 94.2 & 93.5 / 98.5 & 96.4 / \textbf{98.9} & 96.4 / \textbf{98.9}\\
    
    Capsules& 61.4 / 60.7 & 74.7 / 98.5 & 75.1 / 99.1 & 71.2 / 75.3 & 63.2 / 94.7 & 82.9 / 99.4 & \textbf{84.8} / \textbf{99.5}\\
    
    Cashew& \textbf{97.8} / 86.4 & 95.1 / 83.5 & 97.2 / 98.5 & 90.7 / 91.2 & 94.8 / \textbf{99.1} & 96.5 / 98.6 & 96.9 / 98.5\\
    
    Chewinggum& 85.8 / 98.6 & 94.8 / 96.8 & 99.0 / \textbf{98.9} & 91.4 / 98.6 & \textbf{99.1} / 98.5 & 98.1 / \textbf{98.9} & 98.2 / \textbf{98.9}\\
    
    Fryum& 88.6 / 96.7 & \textbf{97.4} / 87.2 & 95.7 / 92.7 & 88.6 / \textbf{97.3} & 93.1 / 95.9 & 95.8 / 92.0 & 96.5 / 92.1\\
    
    Macaroni1& 95.2 / 96.2 & 97.2 / \textbf{99.9} & 97.4 / 99.3 & 98.3 / 97.3 & 88.2 / 98.7 & 98.3 / 99.7 & \textbf{98.5} / 99.7\\
    
    Macaroni2& \textbf{87.9} / 87.5 & 85.0 / \textbf{99.2} & 76.4 / 98.5 & 86.3 / 89.2 & 66.5 / 96.7 & 81.4 / 99.1 & 81.1 / \textbf{99.2}\\
    
    PCB1& 72.1 / 66.9 & 47.6 / 88.7 & 98.0 / 99.5 & 77.4 / 75.2 & 97.0 / 99.1 & 98.7 / \textbf{99.8} & \textbf{98.8} / \textbf{99.8}\\
    
    PCB2& 50.7 / 71.1 & 89.8 / 91.3 & \textbf{97.5} / \textbf{98.7} & 62.2 / 67.3 & 89.4 / 96.6 & 96.4 / 98.3 & 96.6 / 98.3\\
    
    PCB3& 90.5 / 95.1 & 92.0 / 98.0 & \textbf{98.2} / 98.1 & 74.3 / 94.8 & 97.9 / 83.2 & 97.7 / \textbf{99.3} & 97.8 / \textbf{99.3}\\
    
    PCB4& 83.1 / 89.0 & 98.6 / 96.8 & 99.5 / \textbf{98.2} & 80.9 / 89.9 & 98.6 / 98.1 & 99.7 / 98.1 & \textbf{99.8} / \textbf{98.2}\\
    
    Pipe fryum& 81.1 / 81.8 & \textbf{99.8} / 85.8 & 99.7 / 98.7 & 72.2 / 87.3 & 99.1 / \textbf{99.3} & 98.8 / 98.7 & 98.8 / 98.7\\

    \hline
    Total avg.& 82.1 / 85.65 & 88.65 / 93.52 & 94.40 / 98.25 & 81.25 / 88.13 & 90.05 / 96.56 & 95.06 / 98.40 & \textbf{95.35} / \textbf{98.43}\\
    
    \hline
  \end{tabularx}
  \end{center}
  }
\end{table*}

\subsubsection{Implementation Details} We used a WideResNet50~\cite{zagoruyko2016wrn} pre-trained on ImageNet as the feature extractor. 
An inverse WideResNet50 is utilized as the discriminative network, which is operated with up-sampling, concatenating, and dimension compression operations. It finally predicts a feature map with 2 channels with the same dimensions as the input image. For the contrastive learning network, we performed adaptive average pooling operation on the feature maps of each layer and then fed them to the projector and predictor respectively. The projector and predictor consist of a 3-layer and 1-layer MLP respectively. During training, we removed the gradient of only the projector branch, and the projectors of both branches shared the weights.
For the anomaly generator, we experimented with CutPaste~\cite{li2021cutpaste} and Perlin noise~\cite{perlin1985perlin} to generate a variety of pseudo anomalies. For the contrastive learning network, We used the same approach as~\cite{chen2021simsiam} to construct two cropped images for providing positive samples. In building the memory bank, we employed a coreset subsampling percentage of 0.1 to obtain a compact memory bank. 

In the discriminative network pre-training (DNP) training phase, we used the Adam~\cite{kingma2014adam} optimizer with an initial learning rate of 0.0001. The learning rate was dynamically adjusted with a decay factor ($\gamma=0.2$) from 80 epochs to 90 epochs. In the negative-guided contrastive learning (NCL) training phase, we used the Stochastic Gradient Descent optimizer with a weight decay of 0.0001 and a momentum of 0.9, and employed the cosine annealing algorithm to adjust the learning rate. All images were first resized to $256 \times 256$ pixels and then center-cropped to $224 \times 224$ pixels for training and inference. For all classes of the MVTec AD dataset, we trained 100 epochs in the first stage and another 100 epochs in the second stage. For the BTAD dataset, we trained 150 epochs in the second stage. The batch size was set to 16. All of our experiments were conducted on a single NVIDIA RTX3090Ti.

\subsection{Experiment Results and Analysis}
\label{sec:resultsandanalysis}
\subsubsection{Performances on MVTec AD} 

\begin{figure*}[ht]
  \centering
  \includegraphics[width=\linewidth]{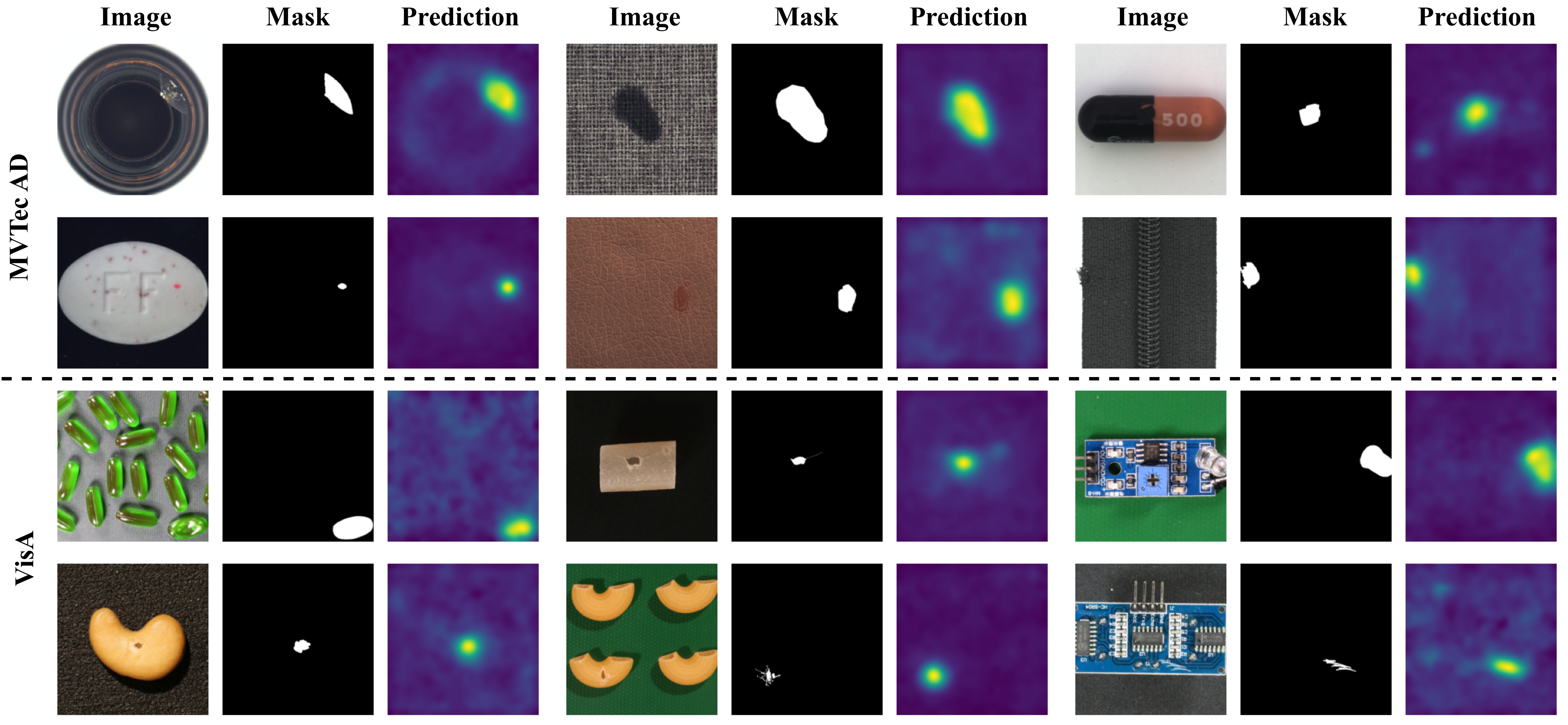}
      \caption{
     Visualization of MVTec AD dataset and VisA dataset. 
     The upper two rows represent sample images, masks, and predictions derived from the MVTec AD dataset, whereas the lower two rows are derived from the VisA dataset.
    }
\label{fig:visualize}
\end{figure*}

We evaluated the anomaly detection performance of our proposed model compared to advanced methods~\cite{cohen2020spade, defard2021padim, roth2022towards, kim2023fapm, deng2022rd4ad, gudovskiy2022cflowad, zhang2023destseg, lei2023pyramidflow, zhang2023mmr}. 
We summarized the texture category including carpet, grid, leather, tile, and wood, while the remaining categories are treated as object categories. We reported image-level and pixel-level AUROC scores for each class, texture, object, and all categories separately, shown in Table~\ref{tab:mvtecad}.
For texture classes, our method achieved 99.08\% and 97.96\% of image- and pixel-level AUROC scores when using CutPaste as the anomaly generator. Similarly, our model achieved competitive performance of 99.16\% and 97.92\% of image-level and pixel-level AUROC scores using Perlin noise as the anomaly generator. 
For object classes, our models obtained 98.69\% and 98.22\% image- and pixel-level AUROC scores, 99.07\% and 98.35\% image-level and pixel-level AUROC scores respectively. 
In summary, our models achieved the best results for at least seven categories in image-level anomaly detection, with five of them reaching 100\%, and produced the best AUROC scores of 99.10\% at the image-level and 98.21\% at the pixel-level compared to advanced methods. We also conducted qualitative visualization to show some representative samples for anomaly localization in the upper two rows of Figure~\ref{fig:visualize}, where we can observe that our models can accurately localize anomalous regions on samples.

\subsubsection{Performances on VisA}
We evaluated our proposed method and the existing methods~\cite{cohen2020spade, zavrtanik2021draem, roth2022towards, yu2021fastflow, gudovskiy2022cflowad} on the VisA dataset, as illustrated in Table~\ref{tab:visa}. Our model achieved 95.06\% image-level AUROC score and 98.40\% pixel-level AUROC score based on CutPaste, while it achieved 95.35\% image-level AUROC score and 98.43\% pixel-level AUROC score based on Perlin noise. Furthermore, when our method employed Perlin noise as a defect generator, it attained the optimal image-level anomaly localization in four subsets and the optimal pixel-level anomaly localization in seven subsets. In the lower two rows of Figure~\ref{fig:visualize}, we additionally visualized samples from select categories of VisA.

\subsubsection{Performances on BTAD} 

We calculated the pixel-level AUROC scores of the three categories and then obtained the average score. We compared our proposed model with existing methods~\cite{cohen2020spade, defard2021padim, roth2022towards, mishra2021vtadl, zheng2022fyd, shin2023spr}.
As shown in Table~\ref{tab:btad}, our model based on CutPaste achieved the best performance of 96.7\% pixel-level AUROC score for product 2 while the localization results. Our models also obtained competitive performance for the other two categories. When using Perlin noise as the anomaly generator, our model achieved the best 97.70\% in average pixel-level AUROC score compared to other methods.

\begin{table}[ht]
  \caption{Pixel-level AUROC (\%) on BTAD dataset. Average results are reported in three categories. The best results are shown in bold.}
  \label{tab:btad}
  \begin{center}
  \begin{tabularx}{\linewidth}{p{0.38\linewidth} | X X X X}
    \toprule
    & 01 & 
    02 & 
    03 & 
    Avg.\\
    \midrule
        SPADE~\cite{cohen2020spade} & 97.3 & 94.4 & 99.1 & 96.93\\
        PaDiM~\cite{defard2021padim} & 96.6 & 95.9 & 98.6 & 97.03\\
        PatchCore~\cite{roth2022towards} & 95.5 & 94.7 & 99.3 & 96.50\\
        VT-ADL~\cite{mishra2021vtadl} & \textbf{99.0} & 94.0 & 77.0 & 90.00\\
        FYD~\cite{zheng2022fyd} & 96.1 & 95.3 & \textbf{99.7} & 97.03\\
        SPR~\cite{shin2023spr} & 96.7 & 96.2 & 95.2 & 96.03\\
    \midrule
        Ours + NSA~\cite{schluter2022nsa} & 97.0 & 96.4 & 99.1 & 97.50\\
        Ours + CutPaste~\cite{li2021cutpaste} & 97.0 & \textbf{96.7} & 99.1 & 97.60\\
        Ours + Perlin~\cite{perlin1985perlin} & 97.4 & 96.6 & 99.1 & \textbf{97.70}\\
    \bottomrule
  \end{tabularx}
  \end{center}
\end{table}

\subsubsection{Ablation Study} 

We explore the optimal configuration of our proposed method. We mainly employed CutPaste as the anomaly generator for the majority of experiments, while we illustrated the anomaly generator we used in our experiments

\textbf{Contrastive learning network}. This study was performed to determine the optimal hierarchy levels for the contrastive learning network. Table~\ref{tab:layers} showed the results of ablation studies with different hierarchy levels when using CutPaste as an anomaly generator. 
We noticed that the model using the features extracted from layer 3 and layer 4 achieved the best performance among all the results, which means that the positive sample comparison learning process requires rich semantic information. When using information from only layer 4, it is insufficient to provide details to reach a good performance. After adding features from layer 2, there was a slight decrease in performance. This suggested that the comparison learning process of the network is more sensitive to shallow features. When using features from a combination of layer 2 and layer 3 without those from layer 4, the results differed from the best result due to less high-level semantic information. In summary, our approach requires rich deep semantic information to foster the capability to represent semantic features.

\begin{table}[ht]
  \caption{Ablation study results on MVTec AD dataset, using different hierarchy levels in NCL training phase. The best results are shown in bold.}
  \label{tab:layers}
  \begin{tabularx}{\linewidth}{p{0.3\linewidth} X X}
    \toprule
    & I-AUROC (\%) & 
    P-AUROC (\%)\\
    \midrule
        Layer 4 & 98.20 & 97.94 \\
        Layer 2+3 & 98.16 & 97.98\\
        Layer 2+4 & 98.16 & 97.95\\
        Layer 2+3+4 & 98.08 & 97.92\\
    \midrule
        Layer 3+4 & \textbf{98.82} & \textbf{98.13}\\
    \bottomrule
  \end{tabularx}
\end{table}

\begin{figure}
    \centering
    \includegraphics[width=\linewidth]{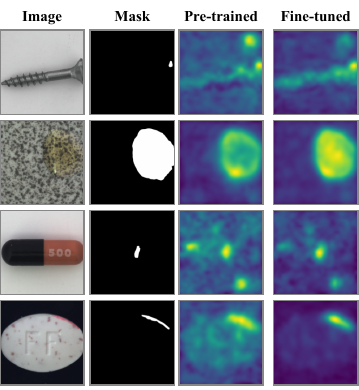}
    \caption{
    Visualization of results with the two-stage training strategy. From left to right, the original image, the ground truth, the result without two-stage training (pre-trained), and the result with two-stage training (fine-tuned).
    }
    \label{fig:compare2}
\end{figure}

\textbf{Two-stage training strategy}. We conducted experiments to investigate the role of discriminative network pre-training (DNP) and negative-guided contrastive learning (NCL) while using Perlin noise as an anomaly generator. All results are shown in Table~\ref{tab:modules}. 
The anomaly detection performance is compared in three scenarios: without DNP and NCL, with NCL only, and with both DNP and NCL. Since the role of DNP can only be manifested on NCL, the case of using only DNP without NCL is not included in the comparison. 
In the absence of DNP and NCL, the proposed network adopted a similar structure to Patchcore~\cite{roth2022towards}, using the pre-trained features for anomaly detection. 
When only NCL was used, the model was equivalent to being trained with only normal augmented samples resulting in unsatisfactory performance due to the lack of explicit negative sample guidance information. By adding both DNP and NCL simultaneously, both negative and positive samples were leveraged to fine-tune the feature extractor, which achieved higher pixel-level localization results. This demonstrated that using negative samples to guide the contrast learning process enables the model to learn feature representations adapted to the normal data samples. As shown in Figure~\ref{fig:compare2}, we also visualized the results of methods using or without DNP and NCL together. We noticed that our model is sensitive to anomalous areas, and the anomalous locations are highlighted prominently on the heat map than the results without two-stage training. This is attributed to the feature extractor learning a robust and generalized feature representation while reducing feature redundancy and domain gap. 

\begin{table}[ht]
  \caption{Ablation study results on MVTec AD dataset, using only NCL (CL-only), with both DNP and NCL. The anomaly generator of the model is based on Perlin noise. The best results are shown in bold.
}
  \label{tab:modules}
  \begin{center}
  \begin{tabularx}{\linewidth}{>{\hsize=.5\hsize}X >{\hsize=.5\hsize}X X X}
    \toprule NCL & DNP
    & I-AUROC (\%) & 
    P-AUROC (\%)\\
    \midrule
        \faTimes & \faTimes & 98.90 & 98.02 \\
        \faCheck & \faTimes & 98.11 & 97.94\\
        \faCheck & \faCheck  & \textbf{99.10} & \textbf{98.21}\\
    \bottomrule
  \end{tabularx}
  \end{center}
\end{table}

\textbf{Loss function}. This study was conducted to explore the effect of different settings on the loss function while utilizing CutPaste as an anomaly generator. Table~\ref{tab:lossfunction} reported the results of using Cross-Entropy (CE) Loss and Focal Loss for different values of the hyperparameter $\lambda$.
We observed that, when $\lambda$ is set to 0.5, a balance was achieved between the positive sample learning process and the negative sample bootstrap learning process, and the results were optimal. And when $\lambda$ was greater or less than 0.5, results showed a small decrease. When set $\lambda=0.5$, the model using Focal loss outperformed the counterpart when using Cross-Entropy Loss.
It can be demonstrated that Focal Loss can fully consider the imbalance between normal and anomalous samples.

\begin{table}[t]
  \caption{Ablation study results of different loss function settings. We compared the results for different values of $\lambda$ and different $L_{neg}$, where CE denotes Cross-Entropy Loss while Focal denotes Focal Loss. The best results are shown in bold.
}
  \label{tab:lossfunction}
  \begin{center}
  \begin{tabularx}{\linewidth}{p{0.08\linewidth} p{0.1\linewidth} p{0.1\linewidth} X X}
    \toprule $\lambda$ & CE & Focal & I-AUROC (\%) & P-AUROC (\%)\\
    \midrule
        \multirow{2}{*}{0.1} & \faCheck & &  98.12 & 97.98 \\
        & & \faCheck &  98.15 & 97.98\\
        \midrule
        \multirow{2}{*}{0.2} & \faCheck & &  98.19 & 97.98 \\ 
        & & \faCheck &  98.04 & 97.96\\
        \midrule
        \multirow{2}{*}{0.5} & \faCheck & & 98.15 & 97.95 \\
         & & \faCheck &  \textbf{98.82} & \textbf{98.13}\\
         \midrule
        \multirow{2}{*}{0.8} & \faCheck & &  98.18 & 97.94 \\
         & & \faCheck &  98.13 & 97.94\\
         \midrule
        \multirow{2}{*}{0.9} & \faCheck & &  98.02 & 97.96 \\
         & & \faCheck &  98.30 & 97.95\\
    \bottomrule
  \end{tabularx}
  \end{center}
\end{table}

\textbf{Synthetic anomaly strategies}. 
We compared three commonly used synthetic defect methods~\cite{li2021cutpaste, schluter2022nsa, perlin1985perlin} as the anomaly generator. These three methods focus on generating random patches, random seamless patches and Perlin noise respectively. As shown in Figure~\ref{tab:anomalygenerator}, our methods incorporating Perlin noise achieved the best performance of 99.10\% image-level AUROC and 98.21\% pixel-level AUROC. We considered that it is mainly attributed to the high randomness of the shape and distribution of the anomalies generated by Perlin noise, enhancing the coarse localization and bootstrap learning ability of the discriminative network.

\begin{table}[ht]
  \caption{Ablation study results on MVTec AD dataset, using different anomaly generator. The best results are shown in bold.
}
  \label{tab:anomalygenerator}
  \begin{center}
  \begin{tabularx}{\linewidth}{X X X}
    \toprule
    Synthetic Method & I-AUROC (\%) & 
    P-AUROC (\%)\\
    \midrule
        CutPaste~\cite{li2021cutpaste} & 98.82 & 98.13 \\
        NSA~\cite{schluter2022nsa} & 98.04 & 97.96\\
        Perlin~\cite{perlin1985perlin} & \textbf{99.10} & \textbf{98.21}\\
    \bottomrule
  \end{tabularx}
  \end{center}
\end{table}

\textbf{Feature fusion in contrastive learning network}. We investigated the effect of different feature fusion operations. As shown in Table~\ref{tab:concatenate}, the model achieved better performance when the concatenation operation was not used. We suggested that the fused features pose challenges in providing an appropriate optimization direction for training multiple layers.

\begin{table}[ht]
  \caption{Ablation study results on MVTec AD dataset, using features with/without concatenation between layer 3 and layer 4. The best results are shown in bold.
}
  \label{tab:concatenate}
  \begin{tabularx}{\linewidth}{X X X}
    \toprule
    Concatenation & I-AUROC (\%) & 
    P-AUROC (\%)\\
    \midrule
        \faCheck & 98.56 & 98.11 \\
        \faTimes & \textbf{98.82} & \textbf{98.13}\\
    \bottomrule
  \end{tabularx}
\end{table}

\textbf{Contrastive learning method}. 
We further conducted experiments to compare two widely used contrastive learning architectures, SimSiam~\cite{chen2021simsiam} and BYOL~\cite{grill2020byol}. The BYOL architecture incorporates additional exponential moving average operations for network updates.
As shown in Table~\ref{tab:clmethod}, the model achieved better performance when using SimSiam, due to its powerful ability to solve the collapsing solution problem~\cite{zhang2022whysimsiam}.

\begin{table}[ht]
  \caption{Ablation study on MVTec AD dataset, using two different contrastive learning methods. The best results are shown in bold.
}
  \label{tab:clmethod}
  \begin{tabularx}{\linewidth}{X X X}
    \toprule
        CL Method & I-AUROC (\%) & 
    P-AUROC (\%)\\
    \midrule
        w/ SimSiam & \textbf{98.82} & \textbf{98.13} \\
        w/ BYOL & 98.45 & 98.08\\
    \bottomrule
  \end{tabularx}
\end{table}

\section{Conclusion}
To bridge the domain gap between the pre-trained and target-specific features for industrial anomaly detection tasks, we proposed a novel two-stage training strategy, named ToCoAD. In the first stage, a discriminative network is trained to coarsely localize anomalies. In the second stage, the pre-trained discriminative network is used to provide negative-guided information, and the contrastive learning network along with the feature extractor are jointly fine-tuned. 
Extensive experiments were conducted to verify the superior performance of our proposed models, achieving 99.10\% image-level AUROC and 98.21\% pixel-level AUROC on the MVTec AD dataset, 95.35\% image-level AUROC and 98.43\% pixel-level AUROC scores on the VisA dataset, and 97.70\% pixel-level AUROC on the BTAD dataset. 
\bibliographystyle{elsarticle-num} 
\bibliography{ref}

\begin{thebibliography}{10}
\expandafter\ifx\csname url\endcsname\relax
  \def\url#1{\texttt{#1}}\fi
\expandafter\ifx\csname urlprefix\endcsname\relax\def\urlprefix{URL }\fi
\expandafter\ifx\csname href\endcsname\relax
  \def\href#1#2{#2} \def\path#1{#1}\fi

\bibitem{tao2022survey22}
X.~Tao, X.~Gong, X.~Zhang, S.~Yan, C.~Adak, Deep learning for unsupervised anomaly localization in industrial images: A survey, IEEE Transactions on Instrumentation and Measurement 71 (2022) 1--21.

\bibitem{liu2024survey2023}
J.~Liu, G.~Xie, J.~Wang, S.~Li, C.~Wang, F.~Zheng, Y.~Jin, Deep industrial image anomaly detection: A survey, Machine Intelligence Research 21~(1) (2024) 104--135.

\bibitem{bergmann2019mvtec}
P.~Bergmann, M.~Fauser, D.~Sattlegger, C.~Steger, Mvtec ad--a comprehensive real-world dataset for unsupervised anomaly detection, in: Proceedings of the IEEE/CVF conference on computer vision and pattern recognition, 2019, pp. 9592--9600.

\bibitem{batzner2024efficientad}
K.~Batzner, L.~Heckler, R.~K{\"o}nig, Efficientad: Accurate visual anomaly detection at millisecond-level latencies, in: Proceedings of the IEEE/CVF Winter Conference on Applications of Computer Vision, 2024, pp. 128--138.

\bibitem{defard2021padim}
T.~Defard, A.~Setkov, A.~Loesch, R.~Audigier, Padim: a patch distribution modeling framework for anomaly detection and localization, in: International Conference on Pattern Recognition, Springer, 2021, pp. 475--489.

\bibitem{roth2022towards}
K.~Roth, L.~Pemula, J.~Zepeda, B.~Sch{\"o}lkopf, T.~Brox, P.~Gehler, Towards total recall in industrial anomaly detection, in: Proceedings of the IEEE/CVF Conference on Computer Vision and Pattern Recognition, 2022, pp. 14318--14328.

\bibitem{cohen2020spade}
N.~Cohen, Y.~Hoshen, Sub-image anomaly detection with deep pyramid correspondences, arXiv preprint arXiv:2005.02357 (2020).

\bibitem{gudovskiy2022cflowad}
D.~Gudovskiy, S.~Ishizaka, K.~Kozuka, Cflow-ad: Real-time unsupervised anomaly detection with localization via conditional normalizing flows, in: Proceedings of the IEEE/CVF winter conference on applications of computer vision, 2022, pp. 98--107.

\bibitem{zavrtanik2021draem}
V.~Zavrtanik, M.~Kristan, D.~Sko{\v{c}}aj, Draem-a discriminatively trained reconstruction embedding for surface anomaly detection, in: Proceedings of the IEEE/CVF International Conference on Computer Vision, 2021, pp. 8330--8339.

\bibitem{schluter2022nsa}
H.~M. Schl{\"u}ter, J.~Tan, B.~Hou, B.~Kainz, Natural synthetic anomalies for self-supervised anomaly detection and localization, in: European Conference on Computer Vision, Springer, 2022, pp. 474--489.

\bibitem{li2021cutpaste}
C.-L. Li, K.~Sohn, J.~Yoon, T.~Pfister, Cutpaste: Self-supervised learning for anomaly detection and localization, in: Proceedings of the IEEE/CVF conference on computer vision and pattern recognition, 2021, pp. 9664--9674.

\bibitem{you2019uda}
K.~You, M.~Long, Z.~Cao, J.~Wang, M.~I. Jordan, Universal domain adaptation, in: Proceedings of the IEEE/CVF conference on computer vision and pattern recognition, 2019, pp. 2720--2729.

\bibitem{nam2021reducingdomaingap}
H.~Nam, H.~Lee, J.~Park, W.~Yoon, D.~Yoo, Reducing domain gap by reducing style bias, in: Proceedings of the IEEE/CVF Conference on Computer Vision and Pattern Recognition, 2021, pp. 8690--8699.

\bibitem{deng2009imagenet}
J.~Deng, W.~Dong, R.~Socher, L.-J. Li, K.~Li, L.~Fei-Fei, Imagenet: A large-scale hierarchical image database, in: 2009 IEEE conference on computer vision and pattern recognition, Ieee, 2009, pp. 248--255.

\bibitem{zavrtanik2021riad}
V.~Zavrtanik, M.~Kristan, D.~Sko{\v{c}}aj, Reconstruction by inpainting for visual anomaly detection, Pattern Recognition 112 (2021) 107706.

\bibitem{pirnay2022intra}
J.~Pirnay, K.~Chai, Inpainting transformer for anomaly detection, in: International Conference on Image Analysis and Processing, Springer, 2022, pp. 394--406.

\bibitem{wyatt2022anoddpm}
J.~Wyatt, A.~Leach, S.~M. Schmon, C.~G. Willcocks, Anoddpm: Anomaly detection with denoising diffusion probabilistic models using simplex noise, in: Proceedings of the IEEE/CVF Conference on Computer Vision and Pattern Recognition, 2022, pp. 650--656.

\bibitem{huang2022caco}
J.~Huang, D.~Guan, A.~Xiao, S.~Lu, L.~Shao, Category contrast for unsupervised domain adaptation in visual tasks, in: Proceedings of the IEEE/CVF conference on computer vision and pattern recognition, 2022, pp. 1203--1214.

\bibitem{thota2021cda}
M.~Thota, G.~Leontidis, Contrastive domain adaptation, in: Proceedings of the IEEE/CVF Conference on computer vision and pattern recognition, 2021, pp. 2209--2218.

\bibitem{dao2023multilabelimageclassification}
S.~D. Dao, H.~Zhao, D.~Phung, J.~Cai, Contrastively enforcing distinctiveness for multi-label image classification, Neurocomputing 555 (2023) 126605.

\bibitem{ki2021weaklysupervisedobjectlocalization}
M.~Ki, Y.~Uh, W.~Lee, H.~Byun, Contrastive and consistent feature learning for weakly supervised object localization and semantic segmentation, Neurocomputing 445 (2021) 244--254.

\bibitem{wang2023dsfwsi}
H.~Wang, E.~Ahn, J.~Kim, A dual-branch self-supervised representation learning framework for tumour segmentation in whole slide images, arXiv preprint arXiv:2303.11019 (2023).

\bibitem{wang2023oneshotorgansegmentation}
B.~Wang, Q.~Li, Z.~You, Self-supervised learning based transformer and convolution hybrid network for one-shot organ segmentation, Neurocomputing 527 (2023) 1--12.

\bibitem{chen2022unpairedderain}
X.~Chen, J.~Pan, K.~Jiang, Y.~Li, Y.~Huang, C.~Kong, L.~Dai, Z.~Fan, Unpaired deep image deraining using dual contrastive learning, in: Proceedings of the IEEE/CVF conference on computer vision and pattern recognition, 2022, pp. 2017--2026.

\bibitem{tian2020explorecl}
Y.~Tian, C.~Sun, B.~Poole, D.~Krishnan, C.~Schmid, P.~Isola, What makes for good views for contrastive learning?, Advances in neural information processing systems 33 (2020) 6827--6839.

\bibitem{kim2020advsslcl}
M.~Kim, J.~Tack, S.~J. Hwang, Adversarial self-supervised contrastive learning, Advances in Neural Information Processing Systems 33 (2020) 2983--2994.

\bibitem{ho2020clae}
C.-H. Ho, N.~Nvasconcelos, Contrastive learning with adversarial examples, Advances in Neural Information Processing Systems 33 (2020) 17081--17093.

\bibitem{cover1967knn}
T.~Cover, P.~Hart, Nearest neighbor pattern classification, IEEE transactions on information theory 13~(1) (1967) 21--27.

\bibitem{he2016resnet}
K.~He, X.~Zhang, S.~Ren, J.~Sun, Deep residual learning for image recognition, in: Proceedings of the IEEE conference on computer vision and pattern recognition, 2016, pp. 770--778.

\bibitem{rezende2015nf}
D.~Rezende, S.~Mohamed, Variational inference with normalizing flows, in: International conference on machine learning, PMLR, 2015, pp. 1530--1538.

\bibitem{lei2023pyramidflow}
J.~Lei, X.~Hu, Y.~Wang, D.~Liu, Pyramidflow: High-resolution defect contrastive localization using pyramid normalizing flow, in: Proceedings of the IEEE/CVF Conference on Computer Vision and Pattern Recognition, 2023, pp. 14143--14152.

\bibitem{ma2024sanf}
W.~Ma, Y.~Li, S.~Lan, W.~Wang, W.~Huang, W.~Zhu, Semantic-aware normalizing flow with feature fusion for image anomaly detection, Neurocomputing 590 (2024) 127728.

\bibitem{dosovitskiy2020vit}
A.~Dosovitskiy, L.~Beyer, A.~Kolesnikov, D.~Weissenborn, X.~Zhai, T.~Unterthiner, M.~Dehghani, M.~Minderer, G.~Heigold, S.~Gelly, et~al., An image is worth 16x16 words: Transformers for image recognition at scale, arXiv preprint arXiv:2010.11929 (2020).

\bibitem{zagoruyko2016wrn}
S.~Zagoruyko, N.~Komodakis, Wide residual networks, arXiv preprint arXiv:1605.07146 (2016).

\bibitem{tan2019efficientnet}
M.~Tan, Q.~Le, Efficientnet: Rethinking model scaling for convolutional neural networks, in: International conference on machine learning, PMLR, 2019, pp. 6105--6114.

\bibitem{zhang2023diffusionad}
H.~Zhang, Z.~Wang, Z.~Wu, Y.-G. Jiang, Diffusionad: Denoising diffusion for anomaly detection, arXiv preprint arXiv:2303.08730 (2023).

\bibitem{ho2020ddpm}
J.~Ho, A.~Jain, P.~Abbeel, Denoising diffusion probabilistic models, Advances in neural information processing systems 33 (2020) 6840--6851.

\bibitem{yang2023memseg}
M.~Yang, P.~Wu, H.~Feng, Memseg: A semi-supervised method for image surface defect detection using differences and commonalities, Engineering Applications of Artificial Intelligence 119 (2023) 105835.

\bibitem{perlin1985perlin}
K.~Perlin, An image synthesizer, ACM Siggraph Computer Graphics 19~(3) (1985) 287--296.

\bibitem{chen2021simsiam}
X.~Chen, K.~He, Exploring simple siamese representation learning, in: Proceedings of the IEEE/CVF conference on computer vision and pattern recognition, 2021, pp. 15750--15758.

\bibitem{grill2020byol}
J.-B. Grill, F.~Strub, F.~Altch{\'e}, C.~Tallec, P.~Richemond, E.~Buchatskaya, C.~Doersch, B.~Avila~Pires, Z.~Guo, M.~Gheshlaghi~Azar, et~al., Bootstrap your own latent-a new approach to self-supervised learning, Advances in neural information processing systems 33 (2020) 21271--21284.

\bibitem{he2020moco}
K.~He, H.~Fan, Y.~Wu, S.~Xie, R.~Girshick, Momentum contrast for unsupervised visual representation learning, in: Proceedings of the IEEE/CVF conference on computer vision and pattern recognition, 2020, pp. 9729--9738.

\bibitem{chen2020simclr}
T.~Chen, S.~Kornblith, M.~Norouzi, G.~Hinton, A simple framework for contrastive learning of visual representations, in: International conference on machine learning, PMLR, 2020, pp. 1597--1607.

\bibitem{sohn2016npairloss}
K.~Sohn, Improved deep metric learning with multi-class n-pair loss objective, Advances in neural information processing systems 29 (2016).

\bibitem{caron2020swav}
M.~Caron, I.~Misra, J.~Mairal, P.~Goyal, P.~Bojanowski, A.~Joulin, Unsupervised learning of visual features by contrasting cluster assignments, Advances in neural information processing systems 33 (2020) 9912--9924.

\bibitem{zhang2022whysimsiam}
C.~Zhang, K.~Zhang, C.~Zhang, T.~X. Pham, C.~D. Yoo, I.~S. Kweon, How does simsiam avoid collapse without negative samples? a unified understanding with self-supervised contrastive learning, arXiv preprint arXiv:2203.16262 (2022).

\bibitem{zhang2022dac}
Z.~Zhang, W.~Chen, H.~Cheng, Z.~Li, S.~Li, L.~Lin, G.~Li, Divide and contrast: Source-free domain adaptation via adaptive contrastive learning, Advances in Neural Information Processing Systems 35 (2022) 5137--5149.

\bibitem{wang2022cdcl}
R.~Wang, Z.~Wu, Z.~Weng, J.~Chen, G.-J. Qi, Y.-G. Jiang, Cross-domain contrastive learning for unsupervised domain adaptation, IEEE Transactions on Multimedia (2022).

\bibitem{liu2023simplenet}
Z.~Liu, Y.~Zhou, Y.~Xu, Z.~Wang, Simplenet: A simple network for image anomaly detection and localization, in: Proceedings of the IEEE/CVF Conference on Computer Vision and Pattern Recognition, 2023, pp. 20402--20411.

\bibitem{cimpoi2014dtd}
M.~Cimpoi, S.~Maji, I.~Kokkinos, S.~Mohamed, A.~Vedaldi, Describing textures in the wild, in: Proceedings of the IEEE conference on computer vision and pattern recognition, 2014, pp. 3606--3613.

\bibitem{lin2017focalloss}
T.-Y. Lin, P.~Goyal, R.~Girshick, K.~He, P.~Doll{\'a}r, Focal loss for dense object detection, in: Proceedings of the IEEE international conference on computer vision, 2017, pp. 2980--2988.

\bibitem{sener2017coreset}
O.~Sener, S.~Savarese, Active learning for convolutional neural networks: A core-set approach, arXiv preprint arXiv:1708.00489 (2017).

\bibitem{kim2023fapm}
D.~Kim, C.~Park, S.~Cho, S.~Lee, Fapm: Fast adaptive patch memory for real-time industrial anomaly detection, in: ICASSP 2023-2023 IEEE International Conference on Acoustics, Speech and Signal Processing (ICASSP), IEEE, 2023, pp. 1--5.

\bibitem{deng2022rd4ad}
H.~Deng, X.~Li, Anomaly detection via reverse distillation from one-class embedding, in: Proceedings of the IEEE/CVF Conference on Computer Vision and Pattern Recognition, 2022, pp. 9737--9746.

\bibitem{zhang2023destseg}
X.~Zhang, S.~Li, X.~Li, P.~Huang, J.~Shan, T.~Chen, Destseg: Segmentation guided denoising student-teacher for anomaly detection, in: Proceedings of the IEEE/CVF Conference on Computer Vision and Pattern Recognition, 2023, pp. 3914--3923.

\bibitem{zhang2023mmr}
Z.~Zhang, Z.~Zhao, X.~Zhang, C.~Sun, X.~Chen, Industrial anomaly detection with domain shift: A real-world dataset and masked multi-scale reconstruction, Computers in Industry 151 (2023) 103990.

\bibitem{zou2022visa}
Y.~Zou, J.~Jeong, L.~Pemula, D.~Zhang, O.~Dabeer, Spot-the-difference self-supervised pre-training for anomaly detection and segmentation, in: European Conference on Computer Vision, Springer, 2022, pp. 392--408.

\bibitem{mishra2021vtadl}
P.~Mishra, R.~Verk, D.~Fornasier, C.~Piciarelli, G.~L. Foresti, Vt-adl: A vision transformer network for image anomaly detection and localization, in: 2021 IEEE 30th International Symposium on Industrial Electronics (ISIE), IEEE, 2021, pp. 01--06.

\bibitem{yu2021fastflow}
J.~Yu, Y.~Zheng, X.~Wang, W.~Li, Y.~Wu, R.~Zhao, L.~Wu, Fastflow: Unsupervised anomaly detection and localization via 2d normalizing flows, arXiv preprint arXiv:2111.07677 (2021).

\bibitem{kingma2014adam}
D.~P. Kingma, J.~Ba, Adam: A method for stochastic optimization, arXiv preprint arXiv:1412.6980 (2014).

\bibitem{zheng2022fyd}
Y.~Zheng, X.~Wang, R.~Deng, T.~Bao, R.~Zhao, L.~Wu, Focus your distribution: Coarse-to-fine non-contrastive learning for anomaly detection and localization, in: 2022 IEEE International Conference on Multimedia and Expo (ICME), IEEE, 2022, pp. 1--6.

\bibitem{shin2023spr}
W.~Shin, J.~Lee, T.~Lee, S.~Lee, J.~P. Yun, Anomaly detection using score-based perturbation resilience, in: Proceedings of the IEEE/CVF International Conference on Computer Vision, 2023, pp. 23372--23382.

\end{thebibliography}




\end{document}